\pdfoutput=1
\documentclass[usenames,dvipsnames]{article}
\usepackage[utf8]{inputenc}
\usepackage{enumitem}
\usepackage{graphicx}
\usepackage[final]{dlde_neurips_2022}
\usepackage{caption}
\usepackage{diagbox}
\usepackage{subcaption}
\usepackage[dvipsnames]{xcolor}
\usepackage{hyperref, amsmath, amssymb}

\title{A PINN Approach to Symbolic Differential Operator Discovery with Sparse Data}
\author{Lena Podina\thanks{equal contribution}, \\
  Cheriton School of Computer Science\\
  University of Waterloo\\
  \And
  Brydon Eastman\samethanks\\
  Department of Applied Mathematics\\
  University of Waterloo\\
  \And Mohammad Kohandel
  \\
  Department of Applied Mathematics\\
  University of Waterloo\\}
\date{July 2022}

\renewcommand{\d}{{\rm d}}
\newcommand{\Int}[1]{\left(#1\setminus\partial #1\right)}
\newcommand*\samethanks[1][\value{footnote}]{\footnotemark[#1]}

\begin{document}

\maketitle


\begin{abstract}

Given ample experimental data from a system governed by differential equations, it is possible to use deep learning techniques to construct the underlying differential operators. In this work we perform symbolic discovery of differential operators in a situation where there is sparse experimental data. This small data regime in machine learning can be made tractable by providing our algorithms with prior information about the underlying dynamics. Physics Informed Neural Networks (PINNs) have been very successful in this regime (reconstructing entire ODE solutions using only a single point or entire PDE solutions with very few measurements of the initial condition). We modify the PINN approach by adding a neural network that learns a representation of unknown hidden terms in the differential equation. The algorithm yields both a surrogate solution to the differential equation and a black-box representation of the hidden terms. These hidden term neural networks can then be converted into symbolic equations using symbolic regression techniques like AI Feynman. In order to achieve convergence of these neural networks, we provide our algorithms with (noisy) measurements of both the initial condition as well as (synthetic) experimental data obtained at later times. We demonstrate strong performance of this approach even when provided with very few measurements of noisy data in both the ODE and PDE regime.
    
\end{abstract}

\section{Introduction}\label{sec:intro}

Mechanistic models, used for modeling real-world processes in biology, chemistry and physics, are in the form of a system of differential equations (DE). For example, in systems biology, glycolytic oscillations in yeast is modelled using a set of coupled DEs~\cite{RUOFF2003179}. In chemistry, the toluene dehydrogenation reaction scheme~\cite{belohlav1997application} is represented using ordinary DEs. In physics, the behaviour of a pendulum with damping is modelled using a second-order ordinary DE~\cite{strogatz2018nonlinear}. For a system of DEs to be useful in modelling a process, initial conditions and parameters (henceforth `DE parameters') specific to the system of DEs need to be estimated. If the correct values for the initial conditions and DE parameters are obtained, the DE can not only be used to interpolate between experimental datapoints, but to predict the future state of a dynamical system.


Machine learning algorithms, for instance neural networks (NN), are particularly helpful in representing unknown quantities in a data-driven way~\cite{belohlav1997application}. NNs with a wide enough hidden layer can be used to approximate any function~\cite{pinkus_1999} by tuning its parameters (henceforth `NN parameters'). Hence, NNs have been used to infer DE parameters or even entire DE models, due to their ability to approximate functions.
Many recent applications use NNs augmented with prior knowledge in order to learn underlying DE models\footnote{The (underlying or true) DE model of a process refers to the DEs that produced the experimental data. }
from data~\cite{transferlearningChakraborty,RAISSI2019686,lu2021neural,rackauckas2020universal,meng2020composite,lu2021deep}. However, acquiring sufficient data to fit these values accurately using NNs is difficult. A method that can function in low-data regimes by leveraging the known structure of the model is needed.

Two prominent NN-based methods that learn DE models from data are physics-informed neural networks (PINN)~\cite{RAISSI2019686} and universal differential equations (UDE)~\cite{rackauckas2020universal}.
In UDEs, each unknown component of the DE model is approximated by a NN and a hard DE constraint is employed. That is, the best-fit DE is satisfied at all times during training. However, UDEs are not robust to noise, require a lot of data, and SINDy, as employed in \cite{rackauckas2020universal}, does not succeed in finding the true mechanistic model reliably.
PINNs assume the form of the true DE and fits its parameters via a soft constraint (relaxing the requirement that the NN should satisfy the best-fit DE exactly), which is added to the NN loss function as an additional loss term referred to as the `PINN loss'.
A drawback of PINNs is that the structure of the DE model must be determined in advance, and there is no way to learn its unknown components using the method as originally proposed. Additionally, as iterative optimization is computationally expensive, PINN loss can fail on stiff DEs~\cite{wang2021understanding}. 

Our approach bridges the limitations of both PINNs (cannot be used when the structure of the DE is not fully known) \cite{RAISSI2019686} and UDEs (not robust to noise and requires lots of data) \cite{rackauckas2020universal}. To address this, we replace the hard constraint of the UDE with that of PINN loss, which allows the approach to learn unknown components of the DE model from data. This approach is robust to noise and performs well in low-data regimes. Additionally, using the AI Feynman algorithm \cite{udrescu2020ai} yielded good results in identifying the underlying hidden terms of the DE model.

In this paper, we claim the following contributions:
\begin{enumerate}[leftmargin=*]
    \item We propose a novel training methodology which combines PINN loss with the UDE framework to allow a PINN-based approach to learn unknown parts of the DE model from data
    \item Our model is robust to noise and learns the unknown DE model components well in the Lotka-Volterra model, cell apoptosis ODE, and Viscous Burgers' equation
    \item Using a symbolic regression algorithm, we reached better identification of the tested systems than in the original UDE paper
\end{enumerate}

\section{Background}\label{sec:background}


As per~\cite{RAISSI2019686}, suppose that the following DE governs a physical process. $u(t,x)$ is an unknown real-valued function of time $(t)$ and position ($x$). Its time derivative is related to its value for each tuple $(t,x)$ with a known function $\mathcal{N}$, and unknown vector of parameters $\theta$. Furthermore, there are $N$ potentially noisy DE measurements $\{t_i,x_i,u_i\}$.

\begin{align}\label{eq:pinn_setup}
    \frac{\partial u(t,x)}{\partial t} = \mathcal{N}[u;\theta], x\in \Omega, \Omega \in R^D, t \in [0,T]
\end{align}

 Although $\theta$ is required in order to find a numerical or analytical function $u$ that satisfies \ref{eq:pinn_setup}, $\theta$ is unknown in this setup. Using the given data, the PINN method from~\cite{RAISSI2019686} can estimate $\theta$ and $u$ simultaneously. Its key component is a neural network $U$, which predicts $u$ given any tuple $(t,x)$. The following loss function is used to train $U$:


\begin{align}\label{eq:loss}
\mathcal{L} = \frac{1}{N} \sum_{i=1}^N|U(t^i,x^i) - u^i| + \frac{1}{M}\sum_{j=1}^M \bigg|\frac{dU}{dt_j} - \mathcal{N}[u;\theta]\bigg|
\end{align}

where $\frac{dU}{dt_j}$ is auto-differentiated through the neural network and evaluated at time $t_j$ and position $x_j$. The first term penalizes $U$ for making predictions that do not match the DE at a predefined set of $M$ collocation points $\{t_j, x_j\}$. The second term penalizes $U$ for making predictions that do not match the data. Note that the second term contains $\theta$, which allows parameter estimate $\hat{\theta}$ to be updated using its gradient with respect to $\mathcal{L}$. At every optimization iteration, the parameters of $U$ are updated along with $\hat{\theta}$.

\vspace{1em}
A UDE~\cite{rackauckas2020universal} is a DE which is defined in part using universal approximators, e.g. a neural network. These NNs can be used to approximate unknown components of $\mathcal{N}$. We will assume that possibly noisy data $\{t_i,x_i,u_i\}$ is available from Eq.~\ref{eq:pinn_setup}. Suppose that $\mathcal{N}$ is a function $g$ composed of $k$ unknown functions $h_i$ and known parameters $\theta$:

\begin{align}\label{eq:ude_N}
\mathcal{N}[u;\theta] = g(u,h_1(u;\theta),\ldots,h_k(u;\theta);\theta)
\end{align}

In~\cite{rackauckas2020universal}, the $h_i$ terms are approximated by a single neural network $H$ with $k$ outputs and fit using iterative optimization such as Adam \cite{adam} or gradient descent~\cite{bishop2006pattern}. Since Eq.~(\ref{eq:pinn_setup}) always holds, the loss function only consists mean squared error between the DE solution and the data. Note that with any particular approximation $H$, the DE (\ref{eq:pinn_setup}) is defined fully and $u$ can be solved for numerically. Hence, the training loop for UDEs involves numerically solving Eq.~(\ref{eq:pinn_setup}), computing the error between the solution and the data, and updating $H$ to better approximate the unknown components of $\mathcal{N}$. At the end of training, $H$ will represent the unknown component of $\mathcal{N}$ and the numerical solution of Eq.~(\ref{eq:pinn_setup}) will yield $u$ that matches the solution of the true DE.
\vspace{1em}

Symbolic regression is a general technique of finding a model that fits data while balancing the simplicity of the model with its accuracy. This problem has been solved with various genetic algorithms (see, among others,~\cite{gaSymbRegression1, gaSymbRegression2}) but since this method is computationally expensive, newer techniques are becoming more popular. For example, AI Feynman~\cite{udrescu2020ai} leverages NNs and symmetry, units, compositionality, etc. and finally returns a list of potential models ranked by error and complexity. Some work ~\cite{SymbolicGPT2021,kamienny2022end} makes use of transformers to find the correct functional form. In our work, we only use them at the final stage after our method has learned an approximate representation of the missing components.





\section{Methods}\label{sec:methods}
Our proposed method is a modification of the PINN approach to discover the functional form of an unknown term within a differential equation. Suppose $\vec{u}(\vec{x}, t)\in\mathbb{R}^m$ for $\vec{x}\in\mathbb{R}^d$. Let $\mathcal{N}$ be a (potentially non-linear) differential operator, then consider time $t$ in the domain $[0, T]\subset\mathbb{R}$ along with a $d$ dimensional, bounded spatial domain $\Omega\subset\mathbb{R}^d$ where $\partial\Omega$ denotes the boundary of $\Omega$. Notably, if $\mathcal{N}$ contains any derivatives, we assume that those derivatives are with respect to the spatial variables only. We then consider problems of the form
\[
\frac{\d}{\d t} \vec{u}(\vec{x}, t) = \mathcal{N}[\vec{u}](\vec{x},t),\quad t\in[0,T],\quad \vec{x}\in\Omega
\]
subject to initial condition
\[
\vec{u}(\vec{x}, 0) = \vec{u}_0(\vec{x}), \quad \vec{x}\in\Omega
\]
and boundary conditions
\[
\beta[\vec{u}](\vec{x}, t) = 0, \quad \vec{x}\in\partial\Omega, \quad t\in [0, T]
\]
where $\beta$ is a (potentially non-linear) differential operator whose derivatives are only with respect to the spatial variables. 

Further, suppose 
\[
\mathcal{N}[\vec{u}](\vec{x}, t) = \mathcal{N}_K[\vec{u}](\vec{x}, t)+ \mathcal{F}[\vec{u}](\vec{x}, t)
\]
where $\mathcal{N}_K$ is some differential operator with known functional form and $\mathcal{F}$ represents some unknown, target differential operator. Similarly, suppose $\beta=\beta_K+\mathcal{B}$ for some known $\beta_K$ and some unknown $\mathcal{B}$.

Finally, one can consider $\Omega=\varnothing$, in which case the underlying differential law is governed by an ordinary differential equation (ODE). In this situation, there is no boundary condition and so no need for $\beta$ (or, equivalently, $\beta$ is the empty function).

Suppose we have $n$ data points $D=\{(t_k, \vec{x}_k, \vec{u}_k)\}_{k=0}^{n-1}$ where $\vec{u}_k=\vec{u}(t_k, \vec{x}_k) + \epsilon_k$ where $\epsilon_k$ is some noise term (potentially $\epsilon_k=0$). We will use this measured data to fit the parameters of (up to) three neural networks. The first network, $F(\vec{u};\theta_F)$, will approximate the target differential operator $\mathcal{F}[\vec{u}]$ by using a neural network with parameters $\theta_F$. The second network, $U(\vec{x}, t; \theta_U)$, will approximate the value of $\vec{u}(x,t)$ by a neural network with parameters $\theta_U$. The third network, $B(\vec{u}; \theta_B)$, will approximate the value of $\mathcal{B}[\vec{u}]$, the unknown target for the boundary condition, with a neural network parameterized by $\theta_B$. To fit these networks, we consider another two sets of collocation points: these sets are $X_P=\{(\vec{x}_k, t_k)\}_{k=0}^{n_P-1} \subset \Int{\Omega}\times(0, T]$ and $X_B=\{(\vec{x}_k, t_k)\}_{k=0}^{n_B-1} \subset (\partial\Omega)\times(0, T]$. 
These sets correspond to locations in the space-time domain where we enforce that our network $U$ satisfies the underlying differential equation (in the case of $X_P$) and the boundary conditions (in the case of $X_B$).

To calculate the gradients for fitting these networks, we consider the loss function 
\[
L(\theta_U, \theta_B, \theta_F)=
    L_M(\theta_U)+
    L_B(\theta_U, \theta_B)+
    L_P(\theta_U, \theta_F).
\]
The first component of the loss is the MSE loss. This loss is the difference in MSE between the measurement value of $\vec{u}\approx \vec{u}_k$ from the input data with the neural network approximation of $\vec{u} \approx U(\vec{x}_k, t_k)$, evaluated at the same space-time location and is given by
\[
L_M(\theta_U)=\frac{1}{n}\sum_{(\vec{x}_k, t_k, \vec{u}_k) \in D}(U(\vec{x}_k, t_k; \theta_U)-\vec{u}_k)^2.
\]

The second component of the loss is the boundary loss. This loss is the mean squared value of the approximated value of the boundary condition and is given by 
\[
L_B(\theta_U, \theta_B)= \frac{1}{n_B}\sum_{(\vec{x}_k, t_k) \in X_{B}}(\beta_K[U](\vec{x}_k, t_k; \theta_U)+B(U(\vec{x}_k, t_k; \theta_U);\theta_B))^2.
\]
The final component of the loss is the PINN loss. This loss is the mean squared error between the value $U_t$, the time derivative of the neural network approximation of $U$, and the value $\mathcal{N}_K[U]+F(U)$.
\[
L_P(\theta_U, \theta_F)=
  \frac{1}{n_P}\sum_{(\vec{x}_k, t_k) \in X_{P}}(\mathcal{N_K}[U](\vec{x}_k, t_k; \theta_U)+F(U(\vec{x}_k, t_k; \theta_U); \theta_F)-U_t(\vec{x}_k, t_k; \theta_U))^2.
\]
This loss function is quite similar to the loss function for PINNs given in \cite{RAISSI2019686}, however here we insert two additional neural networks into the loss function corresponding to the unknown parts of the underlying dynamics in the boundary conditions and the differential equation. To compensate for these additional parameters, we extend the first component of the loss to include more than just initial data (but solution data as well). In this way, $D$ could contain data from the initial condition, data from the boundary, or data from the interior of the domain. 

Practically, one way to select $X_P$ is to simply choose $n_P$ and use Latin hypercube sampling to select $n_P$ points in $\Int{\Omega}\times(0, T]$. A similar construction works for selecting $X_B$. In this way, we are sampling the domain in a space-filling manner.


\section{Results}\label{sec:results}
To demonstrate our approach we trained the above model on the following three test-cases: the Lotka-Volterra equations (an ODE model), a model for cell apoptosis (an ODE model), and the viscous Burgers' equation (a parabolic PDE model).

\subsection{Lotka-Volterra System}\label{sec:lv_results}

We begin our analysis by testing our method on the Lotka-Volterra (LV) model~\cite{berryman1992orgins} of predator-prey interactions. The DE is formulated as follows:
\begin{align*}\label{eq:lotka}
\frac{\d x}{\d t} &= \alpha x - \beta x y \\
\frac{\d y}{\d t} &= - \delta y + \gamma x y ~.
\end{align*}

We take the known portion of the differential equation as $\mathcal{N_K}[U]=[\alpha\,x, -\delta\,y]$ for known parameters $\alpha$ and $\delta$, and seek to learn $F=[F_1, F_2] \approx [-\beta\,x\,y, \gamma\,x\,y]$ from data only, without knowing the target form and without knowing $\beta$ and $\gamma$. 

To generate the synthetic data, $\alpha, \beta, \gamma, \delta$ were fixed at $(1.3, 0.9, 0.8, 1.8)$ respectively, with initial conditions at $(x_0, y_0) = (0.44249296, 4.6280594)$ just as in \cite{rackauckas2020universal}. The time interval was chosen as $[0,3]$ and stayed the same throughout every LV experiment. 

An ODE solver was used to generate data satisfying the LV equations. This yields a set of points $\{t_i, x_i, y_i\}$. Then, Gaussian noise is added to each $x_i$ and $y_i$. Given a particular noise level $\epsilon$, Gaussian noise was added to the data as follows:
\begin{align*}
(x_i)_{noise} &= x_i + \epsilon \cdot \bar{x} \cdot N(0,1)\\
(y_i)_{noise} &= y_i + \epsilon \cdot \bar{y} \cdot N(0,1)
\end{align*}
where $\bar{x}$ denotes the element-wise mean of $x_i$ over all $i$ (similarly for $y$).

First, we demonstrate our approach on noise-free data (Table~\ref{tab:nonoise}) and data with $\epsilon=5\times 10^{-3}$ noise (Table~\ref{tab:noise}) for various values of $n$ (number of data points) and $n_P$ (number of collocation points). We want to show how the hard-to-acquire data can be augmented by taking more collocation points which require no experiments/measurements and come at only the cost of increased computing power. We see that, in contrast to a standard PINN approach, we need to provide more data than just the initial condition. However, even with very sparse measurement data, we can acquire a good discovery by only increasing the number of collocation points. The additional benefit gained from increasing the collocation points is only realized when there is already ample enough experimental data for the algorithm to leverage.

\begin{table}[!htb]
    \begin{minipage}{.5\linewidth}
      \centering
        \begin{tabular}{c|c|c|c|}
     \backslashbox[3.5em]{$n$}{$n_P$} & $10^2$ & $10^3$ & $10^4$ \\\hline
     $1$ & $2\times10^1$ & $2\times10^1$ & $2\times10^1$  \\\hline
     $5$ & $9\times10^{-4}$ & $1\times10^{-3}$ & $9\times10^{-4}$  \\\hline
     $10$ & $2\times10^{-4}$ & $4\times10^{-5}$ & $5\times10^{-6}$   
\end{tabular}
        \caption{Noise Free Data}\label{tab:nonoise}
    \end{minipage}%
    \begin{minipage}{.5\linewidth}
      \centering
        \begin{tabular}{c|c|c|c|}
     \backslashbox[3.5em]{$n$}{$n_P$} & $10^2$ & $10^3$ & $10^4$ \\\hline
     $1$ & $2\times10^{1}$ & $2\times10^{1}$ & $2\times10^{1}$  \\\hline
     $5$ & $6\times10^{-2}$ & $4\times10^{-3}$ & $5\times10^{-3}$  \\\hline
     $10$ & $1\times10^{-3}$& $6\times10^{-4}$ & $8\times10^{-4}$   
\end{tabular}
        \caption{Noisy ($\epsilon=5\times 10^{-3}$) Data}\label{tab:noise}
    \end{minipage} 
    \caption{Tables demonstrating the MSE between $F$ and the true hidden target after training for various values of $n$ and $n_P$.}\label{tab:both}
\end{table}

\begin{figure}[h]
\centering
\begin{subfigure}{.5\textwidth}
  \centering
  \includegraphics[width=1.0\linewidth]{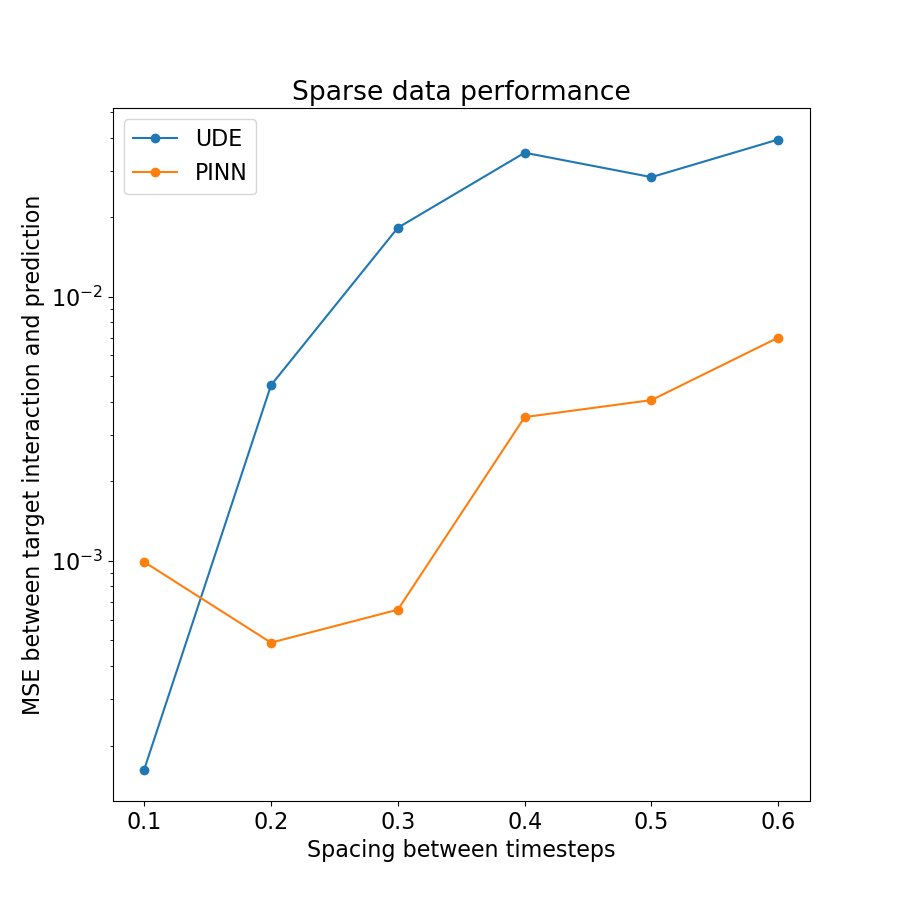}
  \caption{Sparse data}
  \label{fig:comparison1}
\end{subfigure}%
\begin{subfigure}{.5\textwidth}
  \centering
    \includegraphics[width=1.0\linewidth]{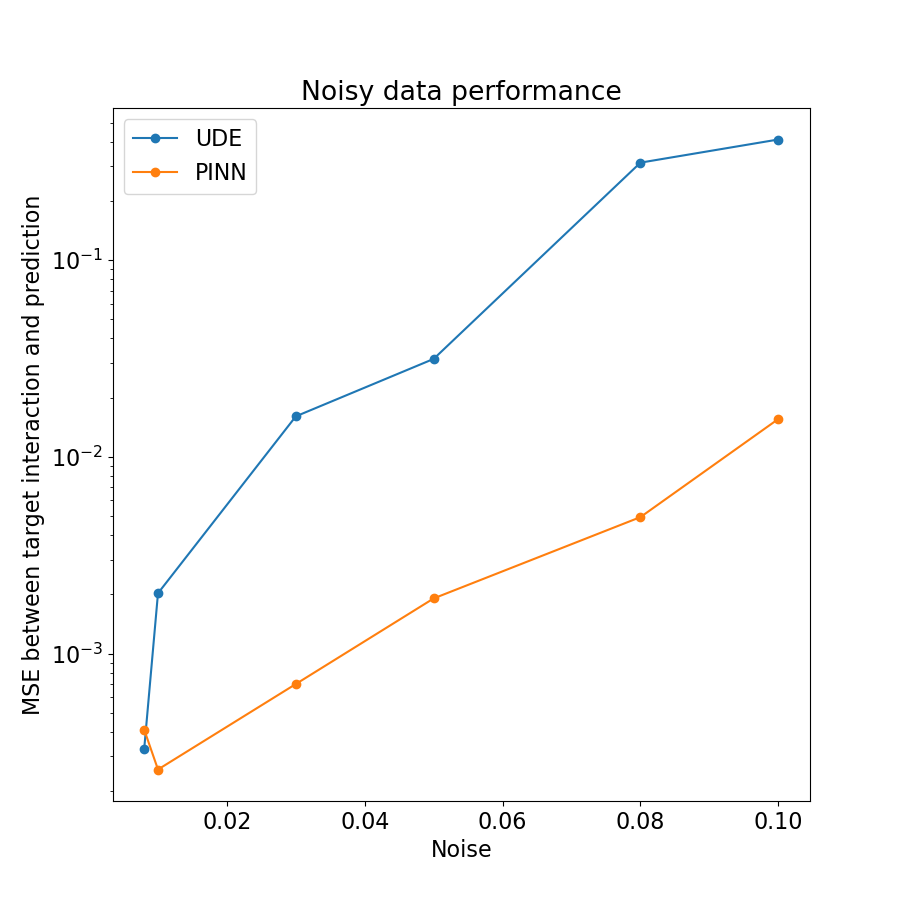}
  \caption{Noisy data}
  \label{fig:comparison2}
\end{subfigure}
\caption{Mean squared error (MSE) of the recovery of the true interaction in comparison between the UDE and the PINN method. The spacing parameter determines how much time passes between datapoints, but the overall time interval $[0,3]$ remains the same. }
\label{fig:test}
\end{figure}

Next, we compare our method's performance to the UDE method. 
We test the two methods on noiseless sparse data (\ref{fig:comparison1}) and on noisy data (Fig \ref{fig:comparison2}). The error is computed as a mean squared error (MSE) taken with respect to the true interaction. At minimal noise level, the UDE approach and PINN approach perform similarly and for the densest data UDEs slightly outperform PINNs. Although increasing either noise or sparsity degrades the performance of both methods, the PINN method consistently attains a lower MSE compared to the UDE method as noise or sparsity increases.

\begin{figure}[h]
\centering
\begin{subfigure}{.5\textwidth}
  \centering
  \includegraphics[width=1.0\linewidth]{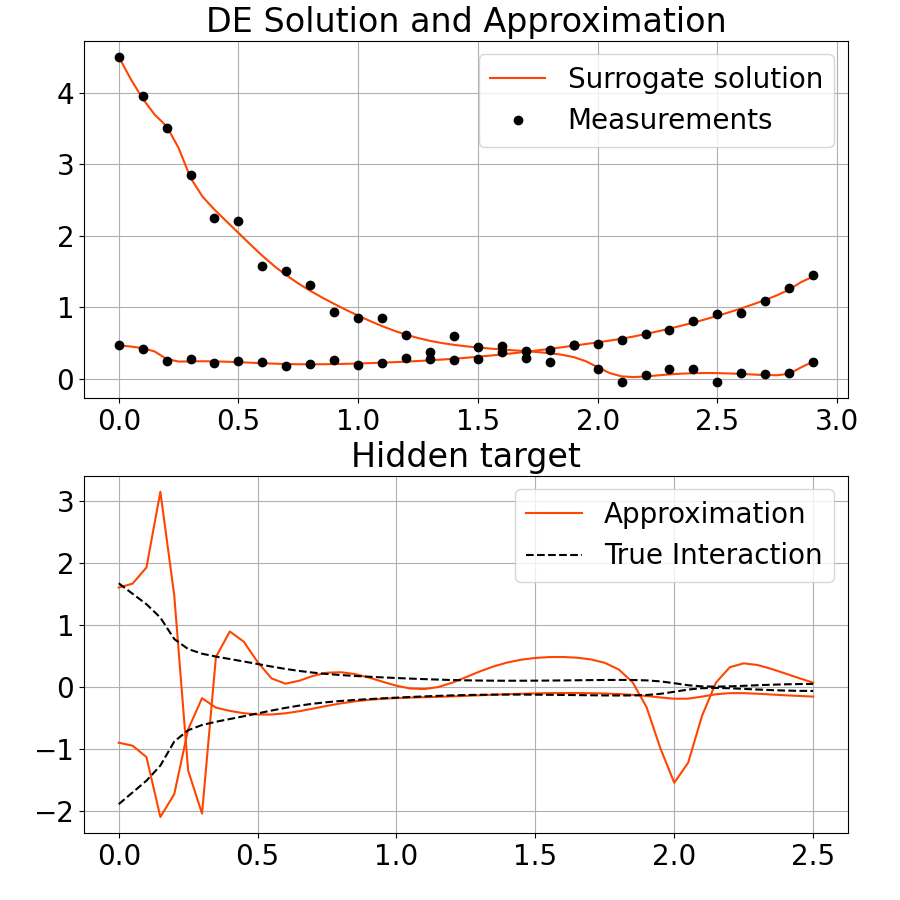}
  \caption{Noisy data}
  \label{fig:noisy_ude}
\end{subfigure}%
\begin{subfigure}{.5\textwidth}
  \centering
  \includegraphics[width=1.0\linewidth]{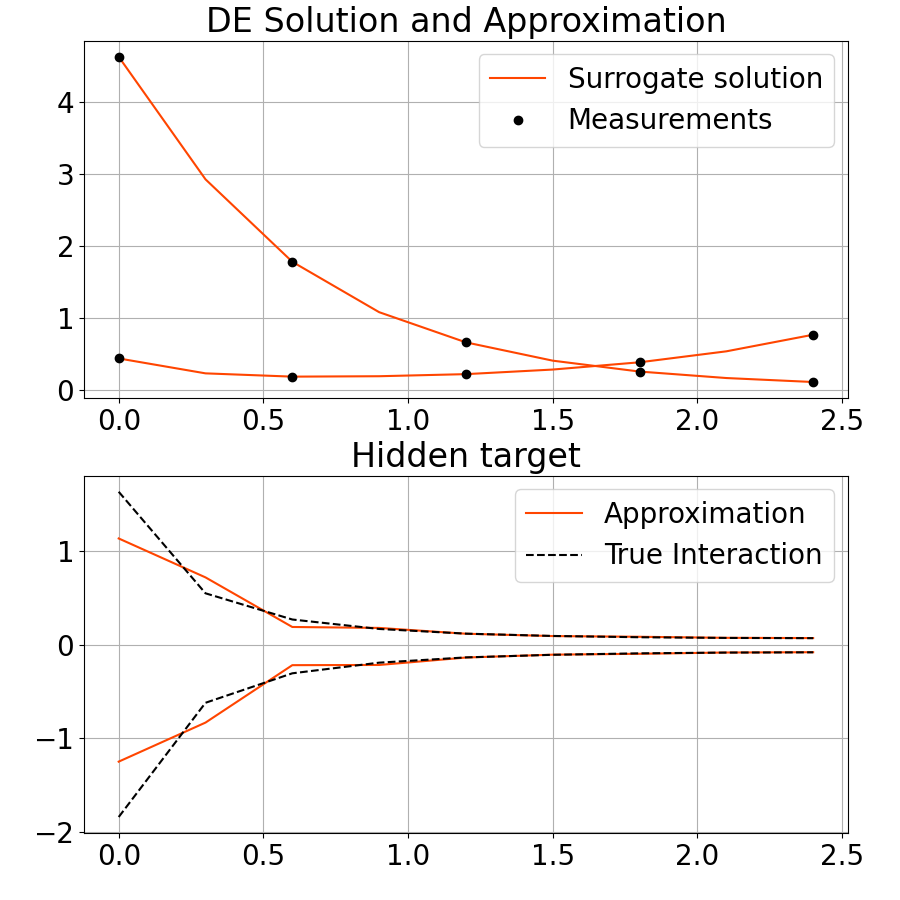}
  \caption{Sparse data}
  \label{fig:sparse_ude}
\end{subfigure}
\caption{UDE method performance}
\label{fig:ude_performance}
\end{figure}

\begin{figure}[h]
\centering
\begin{subfigure}{.5\textwidth}
  \centering
  \includegraphics[width=1.0\linewidth]{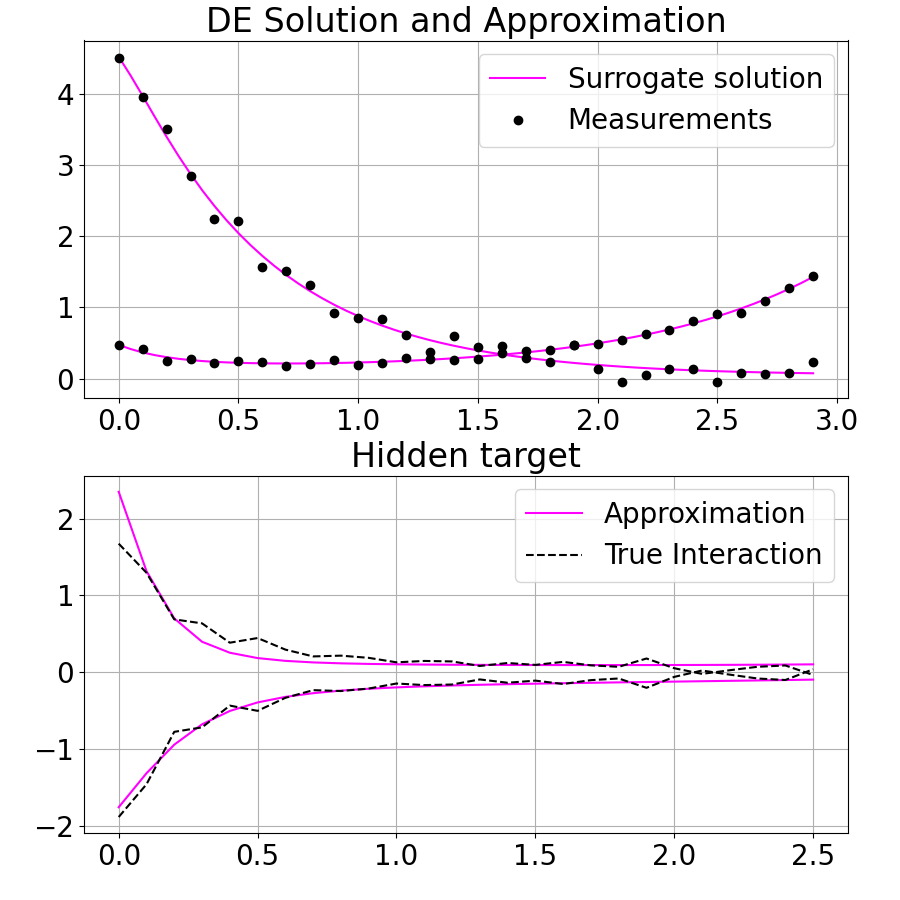}
  \caption{Noisy data}
  \label{fig:noisy_pinn}
\end{subfigure}%
\begin{subfigure}{.5\textwidth}
  \centering
  \includegraphics[width=1.0\linewidth]{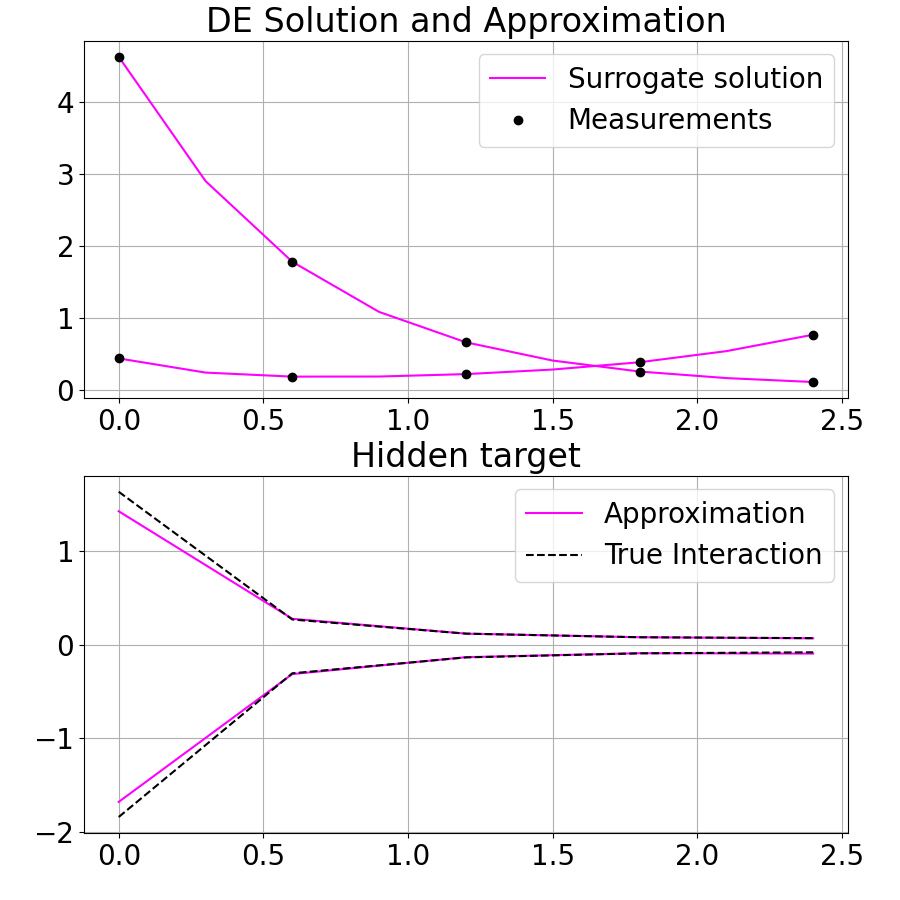}
  \caption{Sparse data}
  \label{fig:sparse_pinn}
\end{subfigure}
\caption{PINN performance}
\label{fig:pinn_performance}
\end{figure}

Figures \ref{fig:ude_performance} and \ref{fig:pinn_performance} show the surrogate solution and hidden terms as recovered by the UDE and PINN methods. The noise level of the noisy data was set at 0.1 and, for the noiseless sparse data, there were 5 points each 0.6 units apart. It is clear that the PINN approach is quite robust to noise and performs well in low-data regimes. The UDE approach performs reasonably on sparse data, but is not robust to noise.

Finally, AI Feynman symbolic regression is run on the neural network output from both our approach and the UDE approach. In all cases, the NNs approximating $F$ were given the training data as input. Then each NN's output was subsequently given to AI Feynman to find the best functional form.
This data is presented in Table~\ref{table:symreg}.
In cases of both sparse and noisy data, AI Feynman correctly recovers the hidden interaction terms more often for our method than it does for the UDE method. If a formula is recovered for both methods, the one recovered for the PINN method is often more accurate.

\begin{table}[h]
\centering
\begin{tabular}{||l||l||l|l||l|l||}
\hline
spacing & noise level & $F_1$ (UDE)     & $F_1$ (PINN)    & $F_2$ (UDE)    & $F_2$ (PINN)   \\ \hline
\hline
0.1     & 0     & \textbf{-0.901 (2.8e-7)} &       --          & \textbf{0.802 (1.1e-6)} & 0.797 (2.5e-6) \\ \hline
0.2     & 0     &      --           &             --              & 0.797 (3.4e-6) & \textbf{0.799 (3.8e-7)} \\ \hline
0.3     & 0     &      --           & \textbf{-0.897 (4e-6)}   &      --          & \textbf{0.798 (1.9e-6)} \\ \hline
0.4     & 0     &      --           & \textbf{-0.888 (8.2e-5)} &      --          & \textbf{0.797 (5.2e-6)} \\ \hline
0.5     & 0     &      --           & \textbf{-0.889 (8.9e-5)} & \textbf{0.760 (1e-3)}   &     --           \\ \hline
0.6     & 0     & -0.892 (4e-3)   & \textbf{-0.890 (1e-5)}   &      --          & \textbf{0.800 (1e-32)}  \\ \hline
\hline
0.1     & 8e-3  & -9.25 (1.8e-3)  & \textbf{-0.906 (1e-5) }          &        --        & \textbf{0.798 (3e-5)}         \\
\hline
0.1     & 1e-2  &        --         & \textbf{-0.911 (3.45e-5)} & \textbf{0.791 (2.3e-5)} & 0.777 (1.5e-4) \\ \hline
0.1     & 3e-2  &       --          & \textbf{-0.960 (1e-3)}   &       --         & \textbf{0.777 (1.5e-4) }\\ \hline
0.1     & 5e-2  &      --           &        --         &       --         & \textbf{0.740 (1.1e-3)} \\ \hline
0.1     & 8e-2  &      --           &        --         &       --         &       --         \\ \hline
0.1     & 1e-1  &       --          &        --        & \textbf{0.887 (2e-3)}   &     --           \\ \hline
\end{tabular}
\caption{Coefficients (with MSE) recovered by AI Feynman from the approximations $F_1$ and $F_2$, comparing over datasets (rows) and method of finding $F_1$ and $F_2$ (columns). True coefficients are  -0.9 for $F_1$ and 0.8 for $F_2$. A dash indicates AI Feynman did not recover the functional form $Cxy$. The best performance is in bold.}
\label{table:symreg}
\end{table}

The terms $\gamma\,x\,y$ and $-\beta\,x\,y$ in the LV equations correspond to the predator's uptake function in the ecological model. This represents the predators' feeding habits as a function of prey population and its resulting effect on both the prey population and the predator's population. The actual form of these functions can take various forms in predator-prey models (see, for instance, \cite{harrison, bolger2020predator}). While we initially modelled this as two unknown, decoupled functions $F_1$ and $F_2$ and learned them independently, we could also have modeled them by a single function with an additional learned parameter as a scaling factor. That is, we could take $F_1=-\phi\,F_2$ and then only explicitly learn $F_2$ and a single parameter $\phi$. This results in regressions that are near identical to the ones presented above, but showcases an important modelling methodology that our method is amenable to and, for more complicated models than LV, may be necessary in order to achieve a high-quality regression.

\subsection{Cell Apoptosis Model}
We also test the method on the Q1 cell apoptosis model from~\cite{wee2006akt}. This is an ODE with three variables, serine-threonine kinase $Akt_s$ (active Akt), $Akt$ (inactive Akt) and tumour suppressor protein $p53$. p53 promotes cell apoptosis, or programmed cell death, and Akt inhibits it. All parameter values were taken from the paper.

\begin{align*}
    v_0 &= k_0 \\
    v_1 &= k_1 \cdot Akt \cdot (j_1 + Akt_s) \\
    v_{m1} &= k_{m1} \cdot Akt_s / (j_{m1} + Akt_s) \\
    v_2 &= k_2 \cdot Akt_s \cdot p53 / (j_2 + p53) \\
    v_{m3} &= k_{m3} \cdot p53 \cdot Akt_s / (j_{m3} + Akt_s) \\
    \frac{dp53}{dt} &= v_0 - v_2 - k_d \cdot p53 \\
    \frac{dAkt_s}{dt} &= v_1 - v_{m1} - v_{m3} \\
    \frac{dAkt}{dt} &= \frac{-dAkt_s}{dt} \\
\end{align*}

\begin{figure}[h]
\centering
\begin{subfigure}{.5\textwidth}
  \centering
  \includegraphics[width=1.0\linewidth]{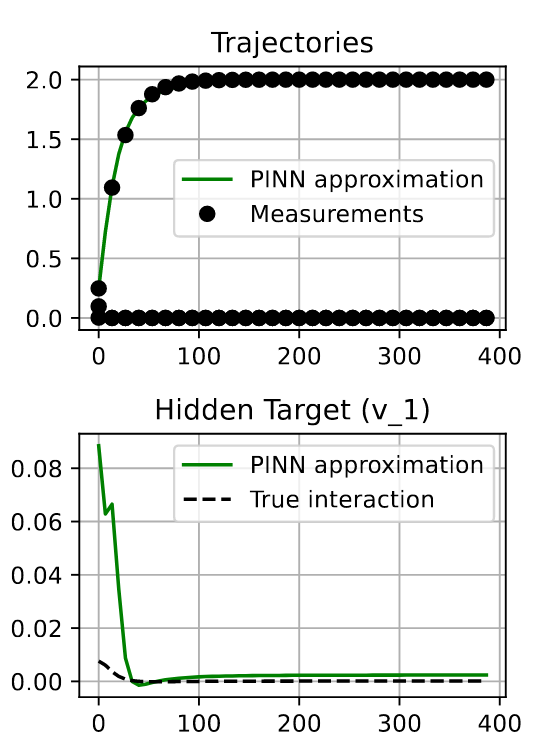}
  \caption{Learning $v_1$}
  \label{fig:v1}
\end{subfigure}%
\begin{subfigure}{.5\textwidth}
  \centering
  \includegraphics[width=1.0\linewidth]{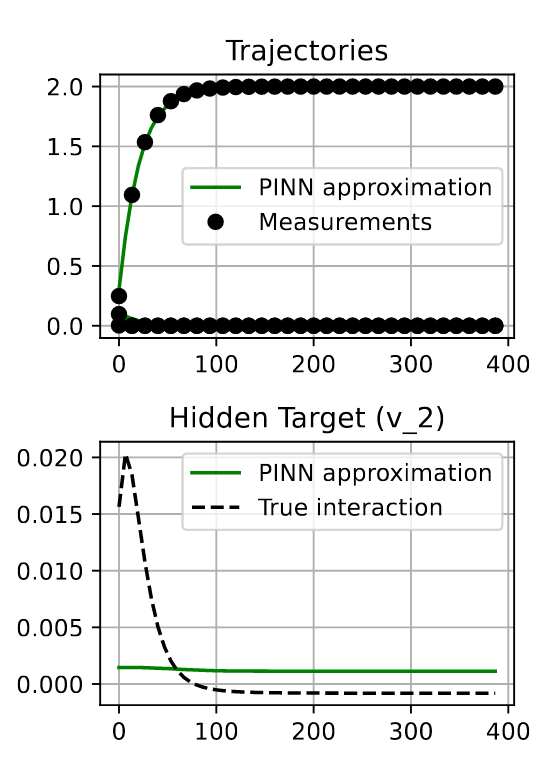}
  \caption{Learning $v_2$}
  \label{fig:v2}
\end{subfigure}
\caption{Learning cell apoptosis components with PINN approach}
\label{fig:pinn_cell}
\end{figure}

With the initial condition $(p53, Akt_s, Akt) = (0.248, 0.0973, 0.0027)$ and 30 noiseless datapoints, both the $v_1$ and $v_2$ interactions were learned to a high degree of accuracy. In Fig \ref{fig:pinn_cell} it can be seen that although the general shape does not match the true interaction 100\%, the mean squared error between the true interaction and the learned function is in fact very small (on the order of $10^{-4}$) and the surrogate solution fits the data very well. As is done in~\cite{yazdani2020systems}, if this method were augmented to handle very small and very large values through scaling, more accurate learning of the interaction would be possible. However, cases arise where the interaction itself is not identifiable and several different interactions learned by $F$ can lead to the minimization of the loss.

\subsection{Viscous Burger's Equation}
\begin{figure}
\centering
\includegraphics[width=0.85\textwidth]{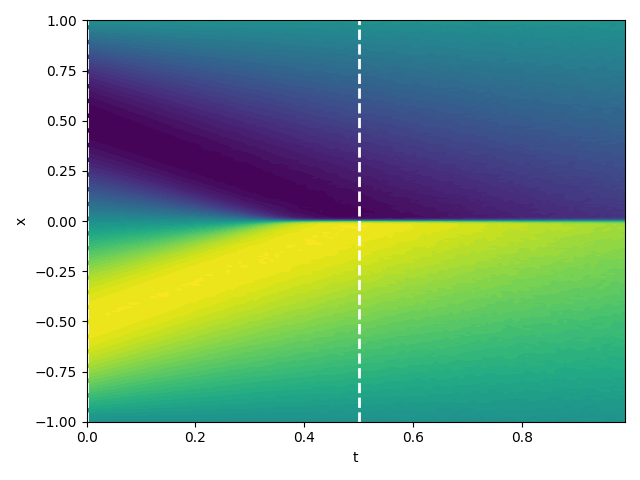}
\caption{The reconstructed solution of Burgers' equation. The two vertical dashed white lines indicate the noisy experimental data that were sampled for the algorithm.}
\label{fig:burgers}
\end{figure}

Finally, our method is easily applied to PDEs (as in the original PINN implementation). Here we present the discovery of both the solution to the PDE where the underlying hidden dynamics of the operator were partially hidden. This reconstruction used only noisy ($\epsilon=5\times10^{-3}$) data obtained from two time points (the initial condition, $t=0$, and a later time at $t=0.5$). While this method can be used to discover the form of the boundary condition as well, here we assume that the homogeneous Dirichlet boundary conditions are known. The PDE in question is
\[
\frac{\partial u}{\partial t}=-u\,\frac{\partial u}{\partial x}+\nu\,\frac{\partial^2 u}{\partial x^2},\quad \nu=\frac{1}{1000\,\pi},\quad u(x,0)=-\sin(\pi\,x)
\]
Here we took $\mathcal{N_K}=\nu\,u_{xx}$ and let the algorithm learn the hidden term $-u\,u_x$. To do this, we gave the $F$ network $u$, $u_x$, and $u_t$ as inputs. This represents an inductive prior where we are assuming that the hidden term depends on first order and lower derivatives of the solution. In our approach, such a prior is necessary (that is, the algorithm cannot learn what order of derivatives to include or not include, it can merely choose which inputs presented to it to utilize). For collocation data we used $n_P=10^4$ and $n_B=10^2$ points sampled from the appropriate parts of the domain $[-1,1]\times[0,1]$ via Latin hypercube sampling. The PDE solution was reconstructed with MSE of $3\times10^{-4}$ and the hidden term was discovered with MSE of $2\times 10^{-2}$. The resulting solution is visualized in Figure~\ref{fig:burgers}.

\section{Conclusion}
In conclusion, our approach is able to recover, with a great degree of accuracy, the symbolic functional form of hidden terms within a differential operator using very sparse measurements of noisy data by utilizing a modification of PINNs. This approach is robust to both noise and sparsity of the data by increasing the number of collocation points (an operation that doesn't require any additional experimentation, just stronger compute capacities). This approach can be applied to discovering the functional form of an unknown ordinary differential equation (ODE) as well as both the functional form of a partial differential operator in a partial differential equation (PDE) and unknown terms in the boundary condition of a PDE. Although PINNs have been noted to perform sub-optimally on stiff equations without modification \cite{stiff1, stiff2}, we have noted promising results in this direction. However, more investigation is needed.

\bibliographystyle{plain}
\bibliography{bibliography}

\begin{thebibliography}{10}

\bibitem{belohlav1997application}
Zdenek Belohlav, Petr Zamostny, Petr Kluson, and Jiri Volf.
\newblock Application of random-search algorithm for regression analysis of
  catalytic hydrogenations.
\newblock {\em The Canadian Journal of Chemical Engineering}, 75(4):735--742,
  1997.

\bibitem{berryman1992orgins}
Alan~A Berryman.
\newblock The orgins and evolution of predator-prey theory.
\newblock {\em Ecology}, 73(5):1530--1535, 1992.

\bibitem{bishop2006pattern}
Christopher~M Bishop and Nasser~M Nasrabadi.
\newblock {\em Pattern recognition and machine learning}, volume~4.
\newblock Springer, 2006.

\bibitem{bolger2020predator}
Tedra Bolger, Brydon Eastman, Madeleine Hill, and Gail Wolkowicz.
\newblock A predator-prey model in the chemostat with holling type ii response
  function.
\newblock {\em Mathematics in Applied Sciences and Engineering}, 1(4):333--354,
  2020.

\bibitem{transferlearningChakraborty}
Souvik Chakraborty.
\newblock Transfer learning based multi-fidelity physics informed deep neural
  network.
\newblock {\em CoRR}, abs/2005.10614, 2020.

\bibitem{harrison}
Gary~W Harrison.
\newblock Global stability of predator-prey interactions.
\newblock {\em Journal of Mathematical Biology}, 8(2):159--171, 1979.

\bibitem{stiff1}
Weiqi Ji, Weilun Qiu, Zhiyu Shi, Shaowu Pan, and Sili Deng.
\newblock Stiff-pinn: Physics-informed neural network for stiff chemical
  kinetics.
\newblock {\em The Journal of Physical Chemistry A}, 125(36):8098--8106, 2021.

\bibitem{kamienny2022end}
Pierre-Alexandre Kamienny, St{\'e}phane d'Ascoli, Guillaume Lample, and
  Fran{\c{c}}ois Charton.
\newblock End-to-end symbolic regression with transformers.
\newblock {\em arXiv preprint arXiv:2204.10532}, 2022.

\bibitem{adam}
Diederik~P Kingma and Jimmy Ba.
\newblock Adam: A method for stochastic optimization.
\newblock {\em arXiv preprint arXiv:1412.6980}, 2014.

\bibitem{lu2021deep}
James Lu, Brendan Bender, Jin~Y Jin, and Yuanfang Guan.
\newblock Deep learning prediction of patient response time course from early
  data via neural-pharmacokinetic/pharmacodynamic modelling.
\newblock {\em Nature machine intelligence}, 3(8):696--704, 2021.

\bibitem{lu2021neural}
James Lu, Kaiwen Deng, Xinyuan Zhang, Gengbo Liu, and Yuanfang Guan.
\newblock Neural-ode for pharmacokinetics modeling and its advantage to
  alternative machine learning models in predicting new dosing regimens.
\newblock {\em Iscience}, 24(7):102804, 2021.

\bibitem{gaSymbRegression1}
Ben McKay, Mark~J Willis, and Geoffrey~W Barton.
\newblock Using a tree structured genetic algorithm to perform symbolic
  regression.
\newblock In {\em First international conference on genetic algorithms in
  engineering systems: innovations and applications}, pages 487--492. IET,
  1995.

\bibitem{meng2020composite}
Xuhui Meng and George~Em Karniadakis.
\newblock A composite neural network that learns from multi-fidelity data:
  Application to function approximation and inverse pde problems.
\newblock {\em Journal of Computational Physics}, 401:109020, 2020.

\bibitem{stiff2}
Christian Moya and Guang Lin.
\newblock Dae-pinn: A physics-informed neural network model for simulating
  differential-algebraic equations with application to power networks.
\newblock {\em arXiv preprint arXiv:2109.04304}, 2021.

\bibitem{pinkus_1999}
Allan Pinkus.
\newblock Approximation theory of the mlp model in neural networks.
\newblock {\em Acta Numerica}, 8:143–195, 1999.

\bibitem{rackauckas2020universal}
Christopher Rackauckas, Yingbo Ma, Julius Martensen, Collin Warner, Kirill
  Zubov, Rohit Supekar, Dominic Skinner, Ali Ramadhan, and Alan Edelman.
\newblock Universal differential equations for scientific machine learning.
\newblock {\em arXiv preprint arXiv:2001.04385}, 2020.

\bibitem{RAISSI2019686}
M.~Raissi, P.~Perdikaris, and G.E. Karniadakis.
\newblock Physics-informed neural networks: A deep learning framework for
  solving forward and inverse problems involving nonlinear partial differential
  equations.
\newblock {\em Journal of Computational Physics}, 378:686--707, 2019.

\bibitem{RUOFF2003179}
Peter Ruoff, Melinda~K Christensen, Jana Wolf, and Reinhart Heinrich.
\newblock Temperature dependency and temperature compensation in a model of
  yeast glycolytic oscillations.
\newblock {\em Biophysical Chemistry}, 106(2):179--192, 2003.

\bibitem{gaSymbRegression2}
Michael Schmidt and Hod Lipson.
\newblock Distilling free-form natural laws from experimental data.
\newblock {\em Science}, 324(5923):81--85, 2009.

\bibitem{strogatz2018nonlinear}
Steven~H Strogatz.
\newblock {\em Nonlinear dynamics and chaos: with applications to physics,
  biology, chemistry, and engineering}.
\newblock CRC press, 2018.

\bibitem{udrescu2020ai}
Silviu-Marian Udrescu and Max Tegmark.
\newblock Ai feynman: A physics-inspired method for symbolic regression.
\newblock {\em Science Advances}, 6(16):eaay2631, 2020.

\bibitem{SymbolicGPT2021}
Mojtaba Valipour, Maysum Panju, Bowen You, and Ali Ghodsi.
\newblock Symbolicgpt: A generative transformer model for symbolic regression.
\newblock In {\em Preprint Arxiv}, 2021.
\newblock Under Review.

\bibitem{wang2021understanding}
Sifan Wang, Yujun Teng, and Paris Perdikaris.
\newblock Understanding and mitigating gradient flow pathologies in
  physics-informed neural networks.
\newblock {\em SIAM Journal on Scientific Computing}, 43(5):A3055--A3081, 2021.

\bibitem{wee2006akt}
Keng~Boon Wee and Baltazar~D Aguda.
\newblock Akt versus p53 in a network of oncogenes and tumor suppressor genes
  regulating cell survival and death.
\newblock {\em Biophysical journal}, 91(3):857--865, 2006.

\bibitem{yazdani2020systems}
Alireza Yazdani, Lu~Lu, Maziar Raissi, and George~Em Karniadakis.
\newblock Systems biology informed deep learning for inferring parameters and
  hidden dynamics.
\newblock {\em PLoS computational biology}, 16(11):e1007575, 2020.

\end{thebibliography}

\end{document}